\documentclass[11pt]{article}

\usepackage[preprint]{acl}
\usepackage{times}
\usepackage{latexsym}
\usepackage[T1]{fontenc}
\usepackage[utf8]{inputenc}
\usepackage{microtype}
\usepackage{inconsolata}
\usepackage{amssymb}
\usepackage{graphicx}
\usepackage{wrapfig}
\usepackage[table]{xcolor}
\usepackage{placeins}
\usepackage{xcolor}
\usepackage{multirow}
\usepackage{booktabs}
\usepackage{diagbox}
\usepackage{makecell}
\usepackage{ragged2e}
\usepackage{colortbl}
\usepackage{xspace}
\usepackage{tabularx}
\usepackage{enumitem}
\usepackage{multicol}
\usepackage{algorithm}
\usepackage{algorithmic}
\usepackage[most]{tcolorbox}
\usepackage{minted}
\usepackage{cleveref}
\usepackage{pifont}

\definecolor{bbblue}{RGB}{235,242,255}
\definecolor{bbgreen}{RGB}{234,248,238}
\definecolor{groupgreen}{RGB}{235,250,230}
\definecolor{groupblue}{RGB}{230,245,255}
\definecolor{groupyellow}{RGB}{255,250,220}
\definecolor{first}{RGB}{253,224,211}
\definecolor{second}{RGB}{222,235,247}
\definecolor{myGreen}{RGB}{34, 139, 34}
\definecolor{myRed}{HTML}{CC3333}
\definecolor{lightblue}{RGB}{230,245,255}

\newcommand{\cmark}{\textcolor{myGreen}{\ding{51}}}
\newcommand{\xmark}{\textcolor{myRed}{\ding{55}}}
\newcommand{\model}{MemAPO\xspace}
\newcommand{\qwenlogo}{\raisebox{-0.2ex}{\includegraphics[height=2.5ex]{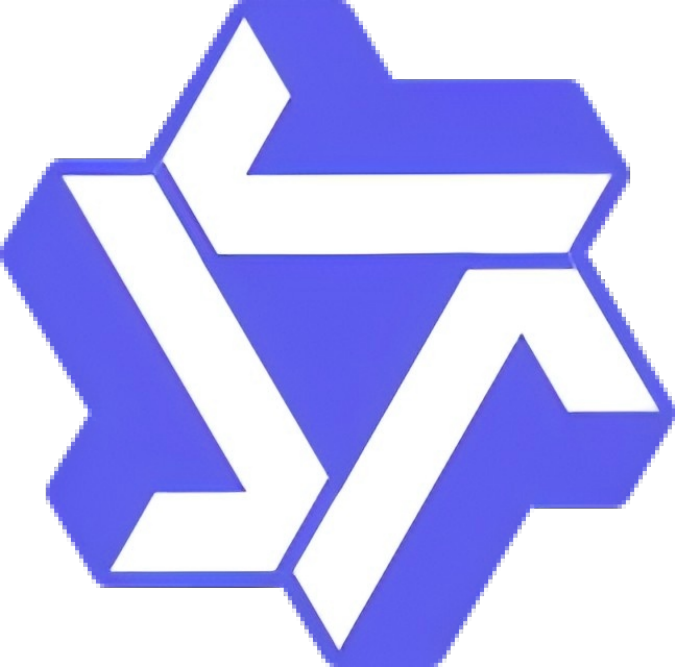}}}
\newcommand{\gptlogo}{\raisebox{-0.7ex}{\includegraphics[height=2.8ex]{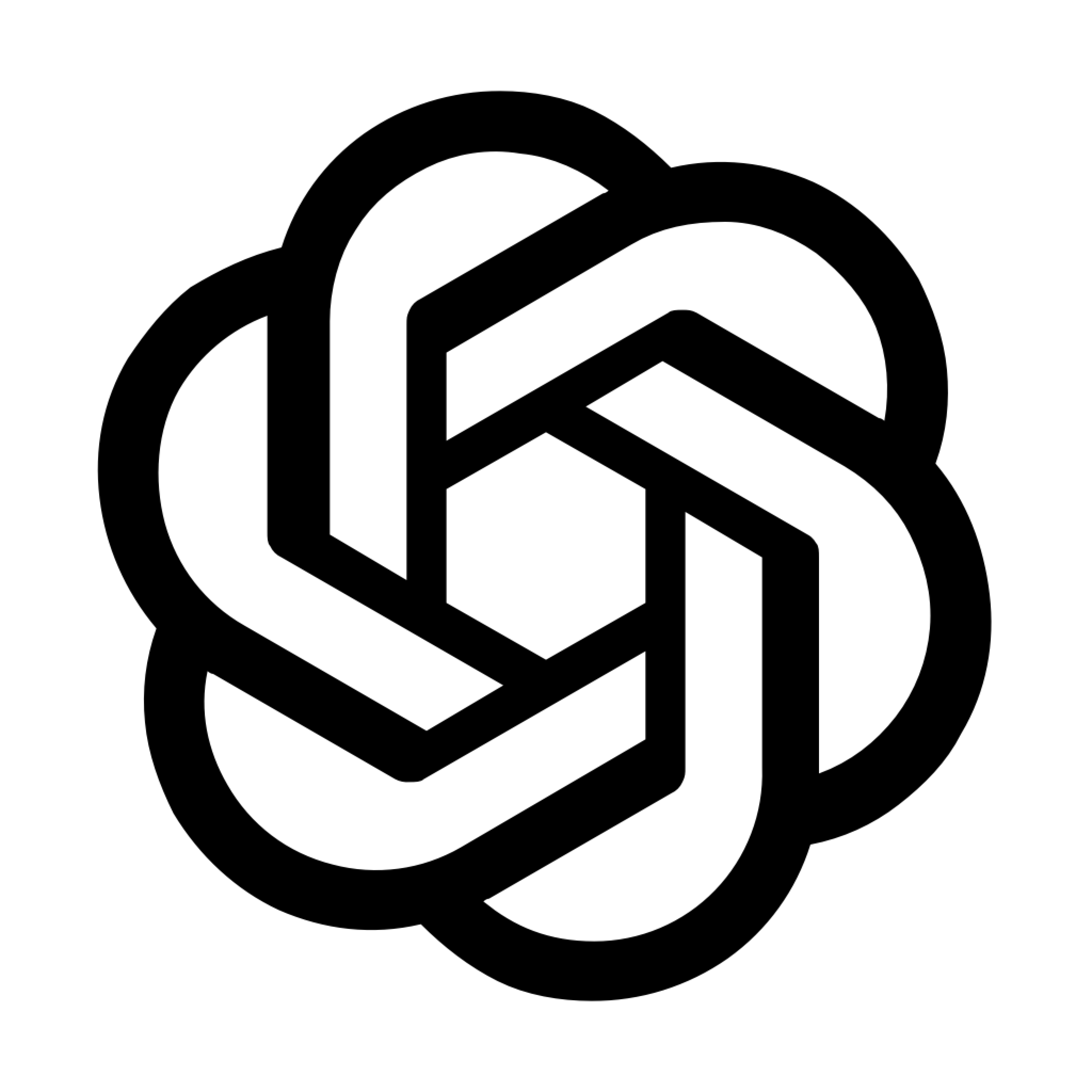}}}
\newcommand{\dblrule}{
  \hline
  \noalign{\vskip 1.5pt}
  \hline
}

\title{Generalizable Self-Evolving Memory for Automatic Prompt Optimization}

\author{Guanbao Liang$^{1^{*}}$,\ Yuanchen Bei$^{2^{*}}$,\ Sheng Zhou$^{1^{\dagger}}$,\ Yuheng Qin$^3$, \\ 
{\bf Huan Zhou}$^3${\bf, Bingxin Jia}$^3${\bf, Bin Li}$^3${\bf,\ Jiajun Bu}$^1$ \\
$^1$Zhejiang University\quad $^2$University of Illinois Urbana-Champaign\quad $^3$Alibaba Group\quad \\
\texttt{\{liangguanbao, zhousheng\_zju, bjj\}@zju.edu.cn, bei4@illinois.edu}\\   
\texttt{\{qinyuheng.qyh, xinyan.zh, bingxin.jbx, yansheng\}@alibaba-inc.com}\\
\small{{* Equal contribution, $\dagger$ Corresponding author}}
}

\begin{document}

\maketitle

\begin{abstract}
  Automatic prompt optimization is a promising approach for adapting large language models (LLMs) to downstream tasks, yet existing methods typically search for a specific prompt specialized to a fixed task. This paradigm limits generalization across heterogeneous queries and prevents models from accumulating reusable prompting knowledge over time. In this paper, we propose \textbf{MemAPO}, a memory-driven framework that reconceptualizes prompt optimization as \textit{\textbf{generalizable and self-evolving experience accumulation}}. MemAPO maintains a dual-memory mechanism that distills successful reasoning trajectories into reusable strategy templates while organizing incorrect generations into structured error patterns that capture recurrent failure modes. Given a new prompt, the framework retrieves both relevant strategies and failure patterns to compose prompts that promote effective reasoning while discouraging known mistakes. Through iterative self-reflection and memory editing, MemAPO continuously updates its memory, enabling prompt optimization to improve over time rather than restarting from scratch for each task. Experiments on diverse benchmarks show that MemAPO consistently outperforms representative prompt optimization baselines while substantially reducing optimization cost.
\end{abstract}

\section{Introduction}
\begin{figure}[t]
  \centering
  \includegraphics[
    width=\columnwidth,
  ]{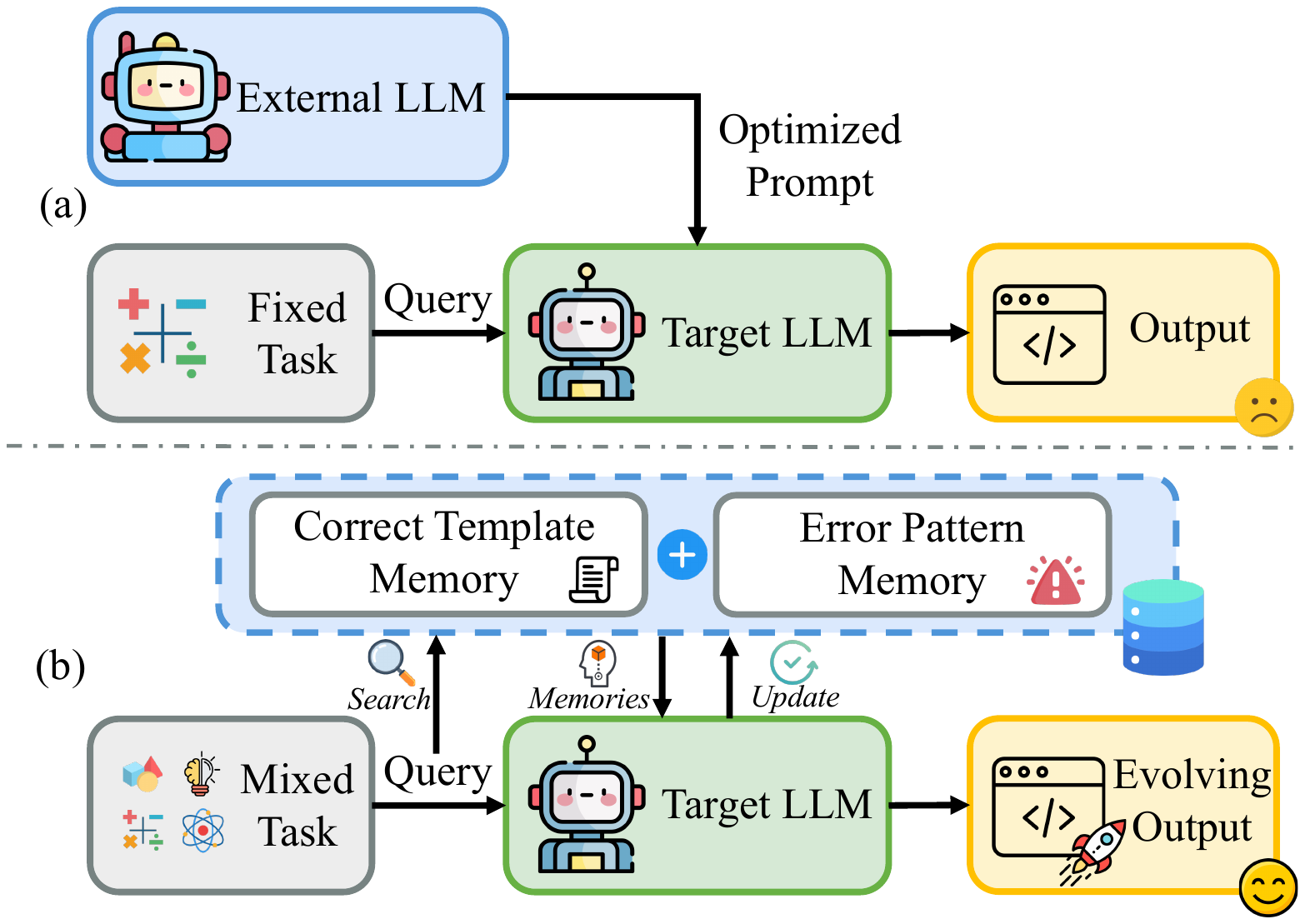}
  \caption{
  \textcolor{black}{Comparison between \model and existing works. (a) Existing paradigm: an external LLM iteratively refines prompts for one specific task. (b) \model: self-organizes reusable memories across multiple tasks.}}
  \label{fig/figure1}
\end{figure}

Large language models (LLMs) have demonstrated remarkable capabilities across a wide range of tasks, including logical reasoning, mathematical problem solving, and knowledge-intensive question answering~\cite{openaio1, gemini, qwen3, deepseekr1, kimik2, glm45}. In practice, however, the performance of LLMs is highly sensitive to prompt design. Carefully crafted prompts can substantially improve model performance, while poorly specified prompts often lead to unstable or suboptimal outputs~\cite{prefixtuning, prompttuning, ptuning, ptuingv2, iclsurvey}. As a result, prompt engineering has become a critical interface for adapting general-purpose LLMs to downstream applications~\cite{wei2022chain, rephrase, huang2026rethinking}.

Despite its importance, designing effective prompts remains a labor-intensive and iterative process that requires both domain expertise and extensive experimentation. To alleviate this burden, recent studies have explored automatic prompt optimization (APO), which leverages LLMs to iteratively refine prompts based on task feedback~\cite{ape, pryzant-etal-2023-automatic, opro, evoprompt}. Existing APO approaches typically treat prompt optimization as a search problem in natural language space. Representative strategies include text-gradient-based optimization~\cite{pryzant-etal-2023-automatic, textgrad}, trajectory-based prompt refinement~\cite{opro, tang2025unleashing}, and evolutionary prompt search~\cite{fernando2023promptbreeder, evoprompt}. These methods have shown promising improvements in controlled settings by automatically generating and refining prompts through iterative evaluation~\cite{textgrad,spo}.

However, most existing APO methods are formulated as a prompt search problem, where the objective is to identify a specific prompt that performs well on a fixed task distribution, as illustrated in Figure~\ref{fig/figure1}-(a). While effective in controlled settings, this formulation faces fundamental challenges in realistic environments.
First, \textbf{prompt optimization is inherently unstable due to the extremely large and stochastic prompt space}. Small variations in prompts can lead to significantly different reasoning behaviors, making optimization brittle and often requiring repeated exploration.
Second, \textbf{useful prompting knowledge is frequently reusable across queries}. Many tasks share underlying reasoning strategies, such as decomposition, verification, or step-by-step deduction. Treating prompt optimization as an isolated search process for each task fails to capture these shared structures and prevents the model from accumulating transferable prompting knowledge.
These observations suggest that prompt optimization should not be viewed merely as prompt search. Instead, it is more naturally framed as a continual knowledge accumulation process, where reusable prompting strategies are progressively discovered and refined through interaction with tasks.
Moreover, simply compressing historical interactions into a single summary is often insufficient. Successful and failed reasoning trajectories contain fundamentally different types of information: successful cases reveal effective reasoning strategies, while failures expose recurring mistakes that should be avoided. A practical system therefore needs to accumulate and organize both types of experience to guide future reasoning.

Motivated by this perspective, as shown in Figure~\ref{fig/figure1}-(b), we propose \textbf{\model}, a self-evolving memory framework for automatic prompt optimization agents. \model enables a target LLM to progressively accumulate prompting experiences and reuse them across heterogeneous tasks. Specifically, \model maintains a dual-memory mechanism that captures both successful strategies and recurrent failure signals. Successful prompting trajectories are distilled into reusable reasoning strategies, while incorrect generations are abstracted into structured failure rules that help prevent recurring mistakes.
To handle heterogeneous query distributions, \model dynamically retrieves relevant experiences from memory based on the characteristics of each input query. Retrieved reasoning strategies provide guidance for effective problem solving, while failure rules help the model avoid previously observed mistakes when constructing prompts. When existing experiences fail to provide adequate guidance, the model autonomously generates new strategies through a self-reflection mechanism and updates the memory accordingly, enabling the prompting knowledge to evolve over time.
Through this process, prompt optimization is transformed \textit{from an externally supervised search procedure into an internal capability that continuously improves through experience accumulation}. We evaluate \model on diverse prompt optimization benchmarks. Experimental results show that \model consistently outperforms representative automatic prompt optimization baselines across multiple datasets and model backbones. In addition, by amortizing optimization cost through experience reuse, \model significantly reduces token consumption and API calls, making it more practical for large-scale deployment.
Our key contributions are summarized as follows:
\begin{itemize}[leftmargin=*]
    \item We introduce a new perspective on prompt optimization by framing it as a self-evolving experience accumulation problem, moving beyond the traditional paradigm of searching for a task-specific optimal prompt.
    \item We propose MemAPO, a self-evolving memory framework that enables LLMs to internalize, store, and reuse prompting experiences through a dual-memory mechanism capturing both strategy templates and error patterns.
    \item Extensive experiments across diverse reasoning benchmarks demonstrate that MemAPO achieves superior performance and generalization while substantially reducing optimization cost.
\end{itemize}

\begin{figure*}[ht]
    \centering
    \includegraphics[width=\textwidth]{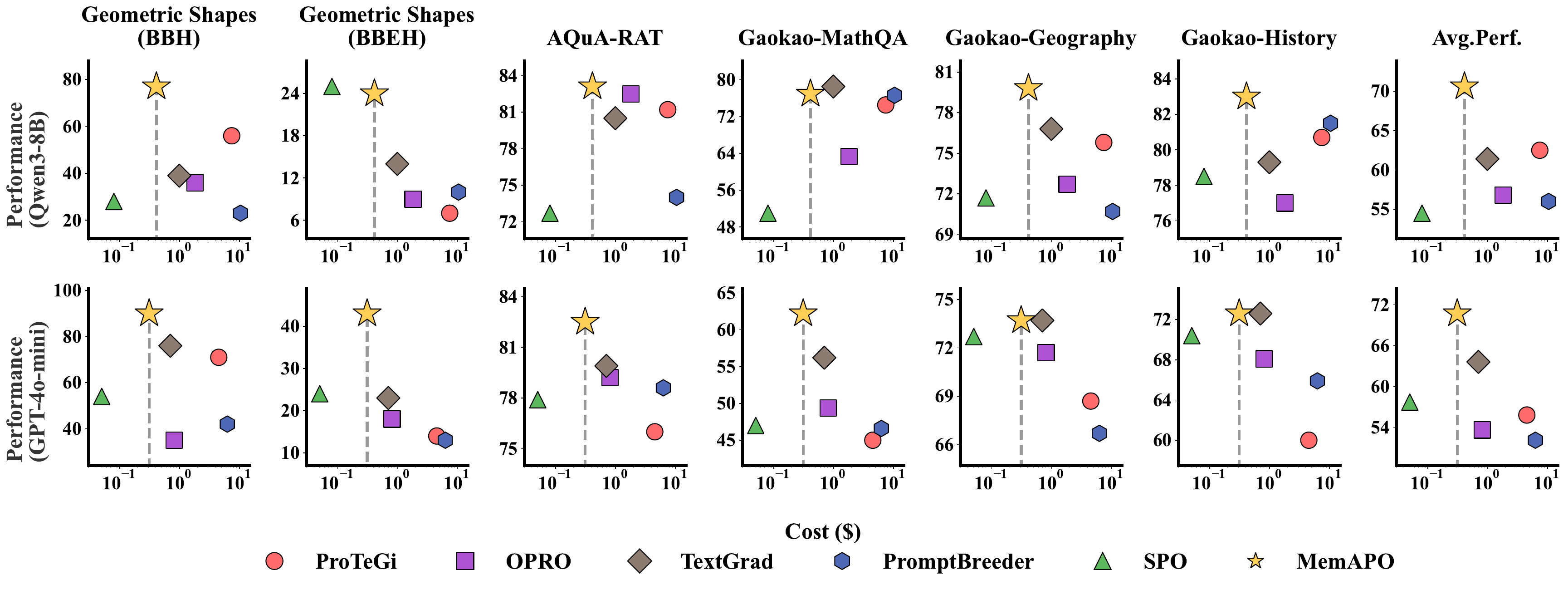}
    \caption{
    Comparison of performance and optimization costs across six automatic prompt optimization methods, six tasks, and two backbones.  
    \model achieves the best average performance across all datasets on both backbones, while notably reducing cost by \textit{approximately 57.2\%} compared to the strong baseline TextGrad.
    }
    \label{fig/cost_performance}
\end{figure*}

\section{Related Work}
\subsection{Automatic Prompt Optimization}
Research on engineering effective prompts for LLMs has been rapidly advancing. Early works like in-context learning \cite{iclsurvey}, chain-of-thoughts \cite{wei2022chain, zeroshotcot}, developed through human insights and extensive experimentation, significantly boost the performance of LLMs. To further automate this process, recent works have shifted towards utilizing LLMs as optimizers, which can be broadly categorized into three paradigms. (1) Text Gradient-Based Optimization \cite{pryzant-etal-2023-automatic, textgrad}, which utilizes natural language feedback as optimization signals to refine prompts; (2) Trajectory-Based Optimization \cite{opro, tang2025unleashing}, which leverages historical prompt-score pairs to guide iterative improvement; and (3) Evolutionary-Based Optimization \cite{evoprompt, fernando2023promptbreeder}, which employs mutation and selection operators on a prompt population. Despite their effectiveness, these approaches generally depend on external, more capable LLM (e.g., GPT-5) serving as the optimizer for a weaker target model (e.g., GPT-4o-mini), which limits their scalability and practical deployment.
 
\subsection{Self-Evolving Agents}
The inherent dependency on external supervision raises a fundamental question: Can a model optimize itself without external guidance? Early works \cite{Self-refine, lu2023self} mainly focus on self-correction and iterative refinement, where a model bootstraps its own performance by critiquing and revising its initial outputs. Drawing inspiration from human cognitive processes, where past experiences are distilled to inform current decisions, recent research \cite{hu2025memory, wei2025evo,bei2026mem} has incorporated memory-augmented mechanisms to store and retrieve historical insights. From the perspective of memory abstraction, existing self-learning methods can be broadly divided into two categories. The first group \cite{reflexion, zhang2023large} focuses on episodic memory, where concrete interaction trajectories and failure cases are stored and reused. In contrast, the second group abstracts historical interactions into higher-level rules or guidelines, as exemplified by \cite{expel, autoguide}. 

\section{Methodology of \model}
\label{sec:method}
\begin{figure*}[ht]
    \centering
    \includegraphics[
    width=\linewidth,
  ]{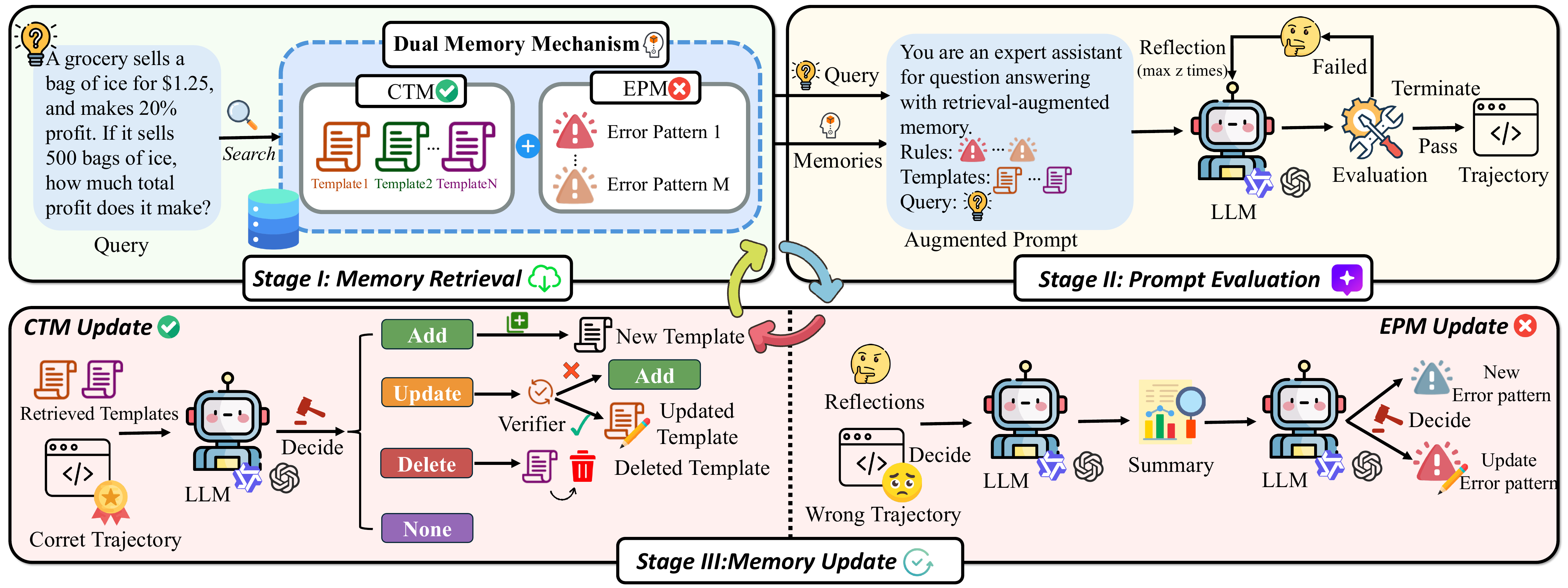}
    \caption{Overall illustration of \model. (I) Memory Retrieval: it first retrieves relevant strategies and failure patterns according to the given query. (II) Prompt Evaluation: Then, the augmented prompt is constructed and iteratively evaluated. (III) Memory Update: Newly acquired experiecnes are consolidated into memory.}
    \vspace{-0.5em}
    \label{fig:main_pipline}
\end{figure*}

\subsection{Problem Definition}
Consider a downstream task $\mathcal{D}=\{(q_i, a_i)\}_{i=1}^{n}$ consisting of $n$ question-answer pairs, where $q_i$ denotes the input query and $a_i$ is the corresponding ground-truth answer. Let $\mathcal{M}_t$ denote the target LLM, and let $\phi(\cdot,\cdot)$ be an evaluation function that measures the quality of a generated response with respect to the ground truth.

The objective of prompt optimization is to learn a prompt $p$ that maximizes the expected task performance of the target model over the data distribution. Formally, the optimal prompt $p^*$ can be defined as
\begin{equation}
p^{*}
= \underset{p}{\arg\max}
\,\mathbb{E}_{(q,a)\sim \mathcal{D}}
\Big[
\phi\big(\mathcal{M}_t(q; p), a\big)
\Big],
\end{equation}
where $\mathcal{M}_t(q; p)$ denotes the response generated by the target model conditioned on the input query $q$ and prompt $p$.
Unlike conventional automatic prompt optimization approaches that search for a single static prompt, we model the prompt as a composition of two components $p = [p_{\mathrm{Inst}},\, p_{\mathrm{Mem}}(q)]$.
The first component $p_{\mathrm{Inst}}$ represents the invariant instruction that specifies general task guidelines such as reasoning style or output format. 
The second component $p_{\mathrm{Mem}}(q)$ is a query-dependent memory augmentation retrieved from historical experience, which provides task-relevant strategies or constraints for the specific query $q$.

Under this formulation, prompt optimization can be viewed as learning how to accumulate, organize, and retrieve reusable prompting knowledge such that the memory-augmented prompt $p$ improves the model's performance across heterogeneous queries.

\subsection{\model Overview}
The overview architecture of \model is illustrated in Figure~\ref{fig:main_pipline}. Specifically, \model maintains a \textbf{Dual-Memory Mechanism} (Section~\ref{sec:dual_memory_mechanism}) that organizes past iteration experiences into two complementary forms of knowledge: (1) \emph{Correct templates} for transferable prompting strategies; and (2) \emph{Error patterns} for preventing repeated mistakes. Based on this memory structure, \model contains three stages:

\textbf{(1) Memory Retrieval} (Section~\ref{sec:memory_retrieval}).  
Given an input query, the framework retrieves relevant strategy templates and global error patterns from memory to construct a memory-augmented prompt that constrains the model's reasoning behavior.

\textbf{(2) Prompt Evaluation} (Section~\ref{sec:prompt_evaluation}).  
The target LLM generates responses conditioned on the retrieved memory, while a self-reflection mechanism iteratively evaluates intermediate outputs and produces corrective feedback to improve reasoning.

\textbf{(3) Memory Update} (Section~\ref{sec:memory_update}).  
Newly acquired experiences are consolidated into memory by either creating/updating templates or summarizing error patterns, enabling the memory repository to evolve over time.

During inference, \model directly retrieves relevant memory and performs a single-pass response generation without invoking the reflection loop, allowing efficient deployment while benefiting from accumulated prompting knowledge.

\subsection{Dual Memory Mechanism}
\label{sec:dual_memory_mechanism}
To enable continual improvement through experience accumulation, \model maintains a structured memory that abstracts reusable knowledge from historical interactions. 
Specifically, \model organizes historical experiences into two complementary memory components: 
\textbf{Correct-Template Memory (CTM)}, which captures reusable successful strategies, and \textbf{Error-Pattern Memory (EPM)}, which summarizes recurring failure modes. 
Together, these two components provide both \textit{positive guidance and negative constraints during prompt construction}.

\paragraph{Correct-Template Memory (CTM).}
Inspired by schema theory in cognitive science, humans interpret new situations by activating previously learned knowledge structures rather than reasoning from scratch~\cite{scheme}.
Similarly, CTM abstracts successful interaction histories into structured templates that can be reused across semantically similar queries.
Formally, the template repository is defined as $\mathcal{E}_{\mathrm{CTM}} = \{T_1, T_2, \dots\}$.
Each template is represented as $T = \{I, S, C\}$.
Here, $I$ denotes the \textbf{index}, which specifies the applicability conditions of the template and serves as a retrieval key. 
$S$ denotes the \textbf{strategy}, which distills transferable reasoning procedures extracted from successful trials. 
By separating strategies from specific inputs, CTM enables knowledge reuse across heterogeneous queries. 
Finally, $C$ denotes the set of supporting \textbf{cases}, which store verifiable exemplars that ground the abstract strategy and facilitate imitation during prompting.

\paragraph{Error-Pattern Memory (EPM).}
While CTM captures successful experiences, Error-Pattern Memory records common failure patterns observed during previous trials. 
Instead of storing raw incorrect responses, EPM abstracts them into high-level reflective rules that describe typical reasoning mistakes and corresponding prevention principles.
Formally, the error-pattern repository is defined as $\mathcal{E}_{\mathrm{EPM}} = \{R_1, R_2, \ldots\}$.
where each $R$ represents a generalized description of a recurrent failure mode derived from reflection on erroneous outputs.

\subsection{Memory Retrieval}
\label{sec:memory_retrieval}
Given a query $q$, the memory retrieval stage selects relevant experiences from the dual-memory repository to construct a memory-augmented prompt.
Formally, the retrieval process can be expressed as
\begin{equation}
\mathcal{T}_q^{(k)}, \mathcal{R}_q
=
\mathcal{F}_{\mathrm{Retrieval}}
\left(
\mathcal{E}_{\mathrm{CTM}},
\mathcal{E}_{\mathrm{EPM}}
\,\middle|\,
q
\right),
\end{equation}
where $\mathcal{T}_q^{(k)}$ denotes the set of retrieved strategy templates from CTM and $\mathcal{R}_q$ represents the retrieved error patterns from EPM.

\paragraph{CTM Retrieval.}
To identify relevant prompting strategies, we retrieve templates from the Correct-Template Memory $\mathcal{E}_{\mathrm{CTM}}$ according to semantic similarity between the query and template indices. 
Specifically, let $\mathbf{e}_q$ denote the embedding of the query and $\mathbf{e}_I$ denote the embedding of a template index $I$. 
The similarity score is computed as $f_{\mathrm{sim}}(q, T) = \cos(\mathbf{e}_q, \mathbf{e}_I)$. The retrieved template set is defined as:
\begin{equation}
\mathcal{T}_q^{(k)}
=
\mathrm{TopK}_{T \in \mathcal{E}_{\mathrm{CTM}}}
\; f_{\mathrm{sim}}(q, T),
\end{equation}
where $\mathcal{T}_q^{(k)}=\{T_q^{(1)}, \ldots, T_q^{(k)}\}$ contains the top-$k$ templates with the highest similarity scores. 
To avoid retrieving irrelevant strategies, we further filter templates whose similarity scores fall below a predefined threshold $\theta_{\mathrm{corr}}$.

\paragraph{EPM Retrieval.}
Unlike CTM, Error-Pattern Memory encodes global behavioral constraints that apply broadly across queries. 
Therefore, all stored error patterns are retrieved and incorporated into the prompt context $\mathcal{R}_q = \mathcal{E}_{\mathrm{EPM}}$.
These patterns provide negative guidance during reasoning by discouraging the model from reproducing previously observed failure modes.

\subsection{Prompt Evaluation}
\label{sec:prompt_evaluation}
Given the retrieved strategy templates $\mathcal{T}_q^{(k)}$ and error patterns $\mathcal{R}_q$, \model constructs a memory-augmented prompt to guide the reasoning process of the target model. 
The augmented prompt is defined as $p_{\mathrm{aug}} = [p_{\mathrm{Inst}}, \mathcal{R}_q, \mathcal{T}_q^{(k)}]$
where $p_{\mathrm{Inst}}$ denotes the task instruction, while $\mathcal{R}_q$ and $\mathcal{T}_q^{(k)}$ provide negative constraints and positive reasoning strategies retrieved from memory.

To acquire informative experiences for future memory updates, \model employs a self-reflection mechanism~\cite{reflexion} that iteratively evaluates intermediate reasoning attempts and produces corrective feedback. 
Let $\mathcal{L}$ denote the set of accumulated reflections, initialized as $\mathcal{L}=\{\emptyset\}$. 
Conditioned on the query, augmented prompt, and current reflections, the target model generates an initial attempt
\begin{equation}
\label{equ:generation}
\hat{a}_0 = \mathcal{M}_t(q, p_{\mathrm{aug}}, \mathcal{L}; p_{\mathrm{ans}}^{\mathrm{meta}}),
\end{equation}
where $p_{\mathrm{ans}}^{\mathrm{meta}}$ is a meta-prompt that instructs the model to incorporate retrieved memories and reflections during reasoning.

The generated answer is then evaluated using the task evaluation function $\phi(\hat{a}_0, a)$. 
If the attempt fails, the model is prompted to reflect on the reasoning process and produce a corrective insight
\begin{equation}
\mathrm{ref} = \mathcal{M}_t(q, p_{\mathrm{aug}}, \mathcal{L}, \hat{a}_0; p_{\mathrm{ref}}^{\mathrm{meta}}),
\end{equation}
where $p_{\mathrm{ref}}^{\mathrm{meta}}$ denotes the meta-prompt for reflection. 
The newly generated reflection is incorporated into the reflection set via $\mathcal{L} \leftarrow \mathcal{L} \cup \{\mathrm{ref}\}$.
The model then retries answer generation using the updated reflection context according to Eq.~\ref{equ:generation}. 
This self-reflection loop continues until either a correct answer is produced or the maximum number of retries is reached. 

\subsection{Memory Update}
\label{sec:memory_update}
After prompt generation and evaluation, \model updates its memory repository based on the outcome of the reasoning attempt. 

\paragraph{CTM Update.}
Correct-Template Memory evolves by incorporating successful reasoning experiences into reusable templates. 
If no template is retrieved for the current query, the model constructs a new template by summarizing the successful interaction. 
Concretely, the target model generates a new template index and strategy using a meta-prompt $p_{\mathrm{corr}}^{\mathrm{meta}}$:
\begin{equation}
(I_{\mathrm{new}}, S_{\mathrm{new}}) =
\mathcal{M}_t(q, \hat{a}, \mathcal{L}; p_{\mathrm{corr}}^{\mathrm{meta}}),
\end{equation}
where $\hat{a}$ denotes the final correct response and $\mathcal{L}$ contains the accumulated reflections. 
The resulting template $T_{\mathrm{new}} = \{I_{\mathrm{new}}, S_{\mathrm{new}}, C_{\mathrm{new}}\}$
is added to the CTM repository $\mathcal{E}_{\mathrm{CTM}}$, where $C_{\mathrm{new}}=\{(q,\hat{a})\}$ stores the corresponding interaction case.

When relevant templates are retrieved, the new successful interaction $(q,\hat{a})$ is appended to their case sets to enrich supporting examples. 
To further maintain template quality, the retrieved templates $\mathcal{T}_q^{(k)}$, together with the interaction $(q,\hat{a})$ and reflections $\mathcal{L}$, are provided to the model via a meta-prompt $p_{\mathrm{act}}^{\mathrm{meta}}$ that determines how the templates should evolve.
The model selects one of the following template operations:
(1) $\mathrm{ADD}$: create a new template when the current interaction represents a novel reasoning pattern.
(2) $\mathrm{UPDATE}$: refine an existing template by incorporating complementary strategy information.
(3) $\mathrm{DELETE}$: remove templates that are contradicted by newly observed evidence.
(4) $\mathrm{NONE}$: retain the existing template without modification. 
For the $\mathrm{UPDATE}$ operation, we perform a verification step: a subset of stored cases from the original template is sampled, and the updated strategy is accepted only if it successfully solves all sampled cases. 
Otherwise, the update is rejected, and a new template is created instead.
Further, to prevent unbounded growth of the template repository, \model periodically performs template consolidation when the number of templates exceeds a predefined capacity. 
Similar templates are merged to maintain a compact and diverse set of strategies.

\paragraph{EPM Update.}
While CTM records successful experiences, Error-Pattern Memory captures recurring failure modes. 
After a failed attempt, the accumulated reflections $\mathcal{L}$ are summarized into a generalized error pattern
\begin{equation}
R_{\mathrm{sum}} =
\mathcal{M}_t(q, a, \mathcal{L}; p_{\mathrm{sum}}^{\mathrm{meta}}),
\end{equation}
where $p_{\mathrm{sum}}^{\mathrm{meta}}$ is a meta-prompt that abstracts the reflection trajectory into a concise rule describing the failure cause.
To maintain a compact error memory, we identify similar error patterns in the repository according to semantic similarity
\begin{equation}
\mathcal{R} =
\{R_i \mid f_{\mathrm{sim}}(R_i, R_{\mathrm{sum}}) > \theta_{\mathrm{error}}\},
\end{equation}
where $\theta_{\mathrm{error}}$ denotes a similarity threshold. 
If similar patterns exist, the model decides whether to refine them by incorporating the new rule; otherwise, a new error pattern is added to the repository. 

\section{Experiment}
\begin{table*}[t]
\centering
\caption{
Main comparison results across three domains.
Tasks from similar domains are merged into a unified training set, and the optimized prompt is evaluated separately on each task.
All methods use the same model for both prompt optimization and evaluation, and experiments are conducted on Qwen3-8B and GPT-4o-mini, respectively.
GeoShape refers to the ``Geometric Shapes" task, Avg. Perf and Avg. Cost refers to the average performance and the average optimization cost, respectively.
}
\vspace{-0.5em}
\label{tab:main_results}
\resizebox{\textwidth}{!}{
\begin{tabular}{c|l|cc|cc|cc|cc}
\Xhline{1.2pt}
\multirow{3}{*}{\textbf{Backbone}} &
\multirow{2}{*}{
\diagbox[
width=18em,
height=3.7em,
innerleftsep=1pt,
innerrightsep=1pt
]{\textbf{Method}}{\textbf{Dataset}}} &
\multicolumn{2}{c|}{\textbf{Logical Reasoning}} &
\multicolumn{2}{c|}{\textbf{Mathematical Calculation}} &
\multicolumn{2}{c|}{\textbf{Knowledge Intensive}} &
\multirow{2}{*}{\textbf{Avg. Perf. $\uparrow$}} &
\multirow{2}{*}{\textbf{Avg. Cost(\$) $\downarrow$}} \\
\cline{3-4}\cline{5-6}\cline{7-8}
& 
& \makecell{\textbf{GeoShape}\\\textbf{(BBH)}}
& \makecell{\textbf{GeoShape}\\\textbf{(BBEH)}}
& \makecell{\textbf{AQuA-RAT}}
& \makecell{\textbf{Gaokao}\\\textbf{MathQA}}
& \makecell{\textbf{Gaokao}\\\textbf{Geography}}
& \makecell{\textbf{Gaokao}\\\textbf{History}}
& & \\
\Xhline{1.2pt}

\multirow{10}{*}{\rotatebox{90}{\makebox[2.8cm][c]{\textbf{Qwen3-8B} \qwenlogo}}}
& IO            & 23.0 &  6.0 & 79.9 & 68.5 & 72.7 & 80.0 & 55.0 & - \\
& CoT \cite{wei2022chain}   & \underline{64.0} &  6.0 & \underline{82.5} & 65.7 & 76.8 & 78.5 & 62.3 & - \\
& Step-Back \cite{stepback}     & 58.0 &  8.0 & 81.8 & 64.5 & 70.7 & 76.3 & 59.9 & - \\
& RaR \cite{rephrase}     & 53.0 &  3.0 & 80.5 & 70.1 & \underline{78.8} & 77.8 & 60.5 & - \\
\cline{2-10}
& ProTeGi \cite{pryzant-etal-2023-automatic}      & 56.0 &  7.0 & 81.2 & 74.5 & 75.8 & 80.7 & \underline{62.5} & 7.44 \\
& OPRO  \cite{opro}    & 36.0 &  9.0 & \underline{82.5} & 63.3 & 72.7 & 77.0 & 56.8 & 1.81 \\
& TextGrad \cite{textgrad}   & 39.0 & 14.0 & 80.5 & \textbf{78.5} & 76.8 & 79.3 & 61.4 & 0.99 \\
& PromptBreeder \cite{fernando2023promptbreeder} & 23.0 & 10.0 & 74.0 & 76.6 & 70.7 & \underline{81.5} & 56.0 & 10.44 \\
& SPO \cite{spo}  & 28.0 & \textbf{25.0} & 72.7 & 51.0 & 71.7 & 78.5 & 54.5 & \textbf{0.08} \\
\cline{2-10}
& \textbf{\model} & \textbf{77.0} & \underline{24.0} & \textbf{83.1} & \underline{76.9} & \textbf{79.8} & \textbf{83.0} & \textbf{70.6} & \underline{0.41} \\

\dblrule

\multirow{10}{*}{\rotatebox{90}{\makebox[2.8cm][c]{\textbf{GPT-4o-mini} \gptlogo}}}
& IO            & 35.0 & 13.0 & 61.7 & 39.4 & 71.7 & \underline{71.9} & 48.8 & - \\
& CoT \cite{wei2022chain}        & 47.0 & 17.0 & 69.5 & 35.5 & 70.7 & 67.4 & 51.2 & - \\
& Step-Back \cite{stepback}       & 61.0 & 25.0 & 59.7 & 41.8 & 62.6 & 60.7 & 51.8 & - \\
& RaR \cite{rephrase}       & 43.0 & \underline{30.0} & 66.9 & 38.6 & 67.7 & 58.5 & 50.8 & - \\
\cline{2-10}
& ProTeGi \cite{pryzant-etal-2023-automatic} & 71.0 & 14.0 & 76.0 & 45.0 & 68.7 & 60.0 & 55.8 & 4.52 \\
& OPRO \cite{opro}               & 35.0 & 18.0 & 79.2 & 49.4 & 71.7 & 68.1 & 53.6 & 0.81 \\
& TextGrad \cite{textgrad}       & \underline{76.0} & 23.0 & \underline{79.9} & \underline{56.2} & \textbf{73.7} & \textbf{72.6} & \underline{63.6} & 0.70 \\
& PromptBreeder \cite{fernando2023promptbreeder} & 42.0 & 13.0 & 78.6 & 46.6 & 66.7 & 65.9 & 52.1 & 6.28 \\
& SPO \cite{spo}                 & 54.0 & 24.0 & 77.9 & 47.0 & \underline{72.7} & 70.4 & 57.7 & \textbf{0.05} \\
\cline{2-10}
& \textbf{\model}  & \textbf{90.0} & \textbf{43.0} & \textbf{82.5} & \textbf{62.2} & \textbf{73.7} & \textbf{72.6} & \textbf{70.7} & \underline{0.31} \\
\Xhline{1.2pt}

\end{tabular}
}
\end{table*}

\subsection{Experimental Setup}
\paragraph{Datasets.}
We evaluate \model on a collection of datasets spanning logical reasoning, mathematical calculation, and knowledge-intensive tasks. For logical reasoning, we adopt two established datasets, \textbf{Big-Bench Hard (BBH)} \cite{bbh} and \textbf{Big-Bench Extra Hard (BBEH)} \cite{bbeh}, and focus on the Geometric Shapes task. We use multiple sub-tasks from \textbf{AGIEval} \cite{zhong2024agieval}, including AQuA-RAT and Gaokao-Math for mathematical calculation, as well as Gaokao-Geography and Gaokao-History for knowledge-intensive reasoning. Following \cite{textgrad}, we sampled portions from original tasks as test sets for each sub-task. Detailed construction procedures are provided in the Appendix ~\ref{appendix:tasks}.
  
\paragraph{Baselines.}
We compare \model with representative APO methods: \textbf{ProTeGi} \cite{pryzant-etal-2023-automatic}, \textbf{OPRO} \cite{opro}, \textbf{TextGrad} \cite{textgrad}, \textbf{PromptBreeder}~\cite{fernando2023promptbreeder}, \textbf{SPO} \cite{spo}. In addition, we also consider conventional prompting methods, comprising \textbf{IO} (direct inference), \textbf{CoT} (chain-of-thought) \cite{wei2022chain}, \textbf{Step-Back} \cite{stepback}, and \textbf{RaR} (rephrase and respond) \cite{rephrase}.

\paragraph{Implementation Details \& Metrics.}
We conduct experiments with GPT-4o-mini and Qwen3-8B \cite{qwen3}, respectively, using the same model as both the task executor and the prompt optimizer. We report the performance using accuracy, following \cite{bbh, bbeh, zhong2024agieval}. We also measure the optimization cost to assess efficiency. 
Other details are provided in the Appendix ~\ref{appendix:baselines}.

\subsection{Experimental Results and Analysis}
\paragraph{Main Results.}
Table \ref{tab:main_results} reports the performance of different APO methods under heterogeneous settings, where tasks from similar domains are jointly optimized and evaluated separately. On GPT-4o-mini, our method achieves the best average performance 70.7\% across all tasks, outperforming the strongest baseline TextGrad by \textbf{7.1\%}. Compared with baselines that exhibit clear trade-offs across tasks, improving some datasets while degrading others, our method consistently yields stable gains across all tasks, demonstrating the effectiveness of the generalizable and self-evolving experience accumulation design. 

To further demonstrate the effectiveness of \model, we conduct experiments on Qwen3-8B backbone and observe similar trends. Specifically, \model achieves 70.6\% average accuracy, exhibiting the best performance on most datasets. These results indicate that our proposed \model generalizes well across both proprietary and open-source models.

\paragraph{Cost Analysis.}
Besides the performance, we present a comparison of optimization costs and performance between {\model} and other methods in Figure \ref{fig/cost_performance}. It shows optimization costs on the x-axis and performance on the y-axis, with methods towards the top-left being more efficient and powerful. Whether using Qwen3-8B or GPT-4o-mini as the target LLM, our method achieves superior performance while ranking second in optimization efficiency across all methods. Compared to the strong baseline TextGrad, it reduces costs by 58.6\% and 55.7\%, respectively. This significant optimization cost reduction can be attributed to the efficient experience reuse, making {\model} suitable for practical applications.

\paragraph{Generalization Analysis.}
\begin{table}[t]
\centering
\caption{Cross-domain (Gaokao MathQA and Gaokao History) experiment using GPT-4o-mini.}
\label{tab:generalization}
\resizebox{0.9\linewidth}{!}{
\begin{tabular}{l|cc}
\hline
\textbf{Method} & \textbf{Gaokao MathQA} & \textbf{Gaokao History} \\
\hline
IO            & 39.4 & \underline{71.9} \\
CoT           & 35.5 & 67.4 \\
StepBack      & 41.8 & 60.7 \\
Rephrase      & 38.6 & 58.5 \\
\hline
ProTeGi       & 45.0 & 53.3 \\
OPRO          & 56.2 & 67.4 \\
TextGrad      & \underline{59.0} & 69.6 \\
PromptBreeder & 37.5 & 71.1 \\
SPO           & 35.9 & \underline{71.9} \\
\hline
\textbf{\model}          & \textbf{61.8} & \textbf{72.6} \\
\hline
\end{tabular}
}
\end{table}
We have discussed the results under an intra-domain setting where tasks from the same domain are jointly optimized. A more challenging and practical scenario arises when mixing tasks from different domains. We construct a heterogeneous dataset by selecting Gaokao Math QA and Gaokao History from the mathematical calculating and knowledge-intensive domains, respectively. As shown in Table ~\ref{tab:generalization}, most baselines suffer from severe performance degradation, exemplified by SPO, whose performance drops 11.1\% on Gaokao MathQA. This indicates that a specific optimized prompt struggles to reconcile the conflicting reasoning demands of different domains. In contrast, \model maintains improvements on two tasks, achieving the best performance, which demonstrates the generalization ability.

\subsection{Ablation Study}
\begin{table}[t]
\centering
\caption{Ablation study on the effect of dual memory components in \model using GPT-4o-mini.}
\resizebox{\columnwidth}{!}{
\begin{tabular}{cc|cc}
\hline
\makecell{\textbf{Correct Template} \\ \textbf{Memory}} &
\makecell{\textbf{Error Pattern}\\ \textbf{Memory}} &
\textbf{AQuA-RAT} &
\makecell{\textbf{Gaokao MathQA}} \\
\hline
\xmark & \xmark & 61.7 & 39.4 \\
\xmark & \cmark & 77.9 & 60.2 \\
\cmark & \xmark & \underline{80.5} & \underline{61.0} \\
\cmark & \cmark & \textbf{82.5} & \textbf{62.2} \\
\hline
\end{tabular}
}
\label{tab:abl_component}
\end{table}

\begin{table}[t!]
\centering
\caption{Ablation study on the effect of self-evolving paradigm. We compare performance across different LLMs for prompt optimization while adopting GPT-4o-mini for prompt evaluation.}
\label{tab:abl_self_evolving}
\resizebox{1.0\linewidth}{!}{
\begin{tabular}{l|cc|cc}
\hline
\multirow{2}{*}{\textbf{Method}}
& \multicolumn{2}{c|}{\textbf{AQUA-RAT}} 
& \multicolumn{2}{c}{\textbf{Gaokao MathQA}} \\
\cline{2-5}
& \textbf{GPT-4o-mini} & \textbf{GPT-5-Chat} 
& \textbf{GPT-4o-mini} & \textbf{GPT-5-Chat} \\
\hline
\textbf{ProTeGi}  & 76.0 & \underline{82.5} & 45.0 & 50.6 \\
\textbf{OPRO}  & 79.2 & 79.2 & 49.4 & 47.0 \\
\textbf{TextGrad} & 79.9 & 81.8 & 56.2 & 60.2 \\
\textbf{PromptBreeder} & 78.6 & 80.5 & 46.6 & 36.3 \\
\textbf{SPO}      & 77.9 & 79.2 & 47.0 & \underline{63.3} \\
\textbf{\model}   & \underline{82.5} & \textbf{83.8} & 62.2 & \textbf{64.5} \\
\hline
\end{tabular}
}
\end{table}

\paragraph{The Effect of Dual Memory Components.} 
To verify the effectiveness of two key components, CTM and EPM, in our dual-memory mechanism, we conduct an ablation experiment on the mathematical domain using GPT-4o-mini, as shown in Table \ref{tab:abl_component}. The experiment results demonstrate that CTM improves the performance by 18.8 and 21.6 on AQuA-RAT and Gaokao MathQA, respectively, while EPM yields improvements of 16.2 and 20.8. Combining with both CTM and EPM, \textbf{\model} obtains further 2.0 and 1.2 gains, respectively, confirming that relevant strategies and failure patterns promote effective reasoning while discouraging known mistakes.

\paragraph{The Effect of Self-Evolving Paradigm.}
\label{sec:self_improving}
We further conduct an ablation study to validate the effectiveness of the self-evolving paradigm on the mathematical domain, as shown in Table ~\ref{tab:abl_self_evolving}. The experiment results show that introducing a stronger LLM as optimizer generally leads to performance improvements (9/12), but may sometimes have the opposite effect, as exemplified by PromptBreeder and OPRO, which exhibit performance degradation on Gaokao MathQA (46.6$\rightarrow$36.3, 49.4$\rightarrow4$7.0). For \model, it still maintains the averaged second-best performance when using GPT-4o-mini as the optimizer, achieving 98.4\% and 96.4\% of the performance obtained with GPT-5, respectively, demonstrating the effectiveness of the self-evolving paradigm designed in \model.

\paragraph{The Impact of Template Number.} 
\begin{figure}[t]
  \centering
  \includegraphics[width=1\columnwidth]{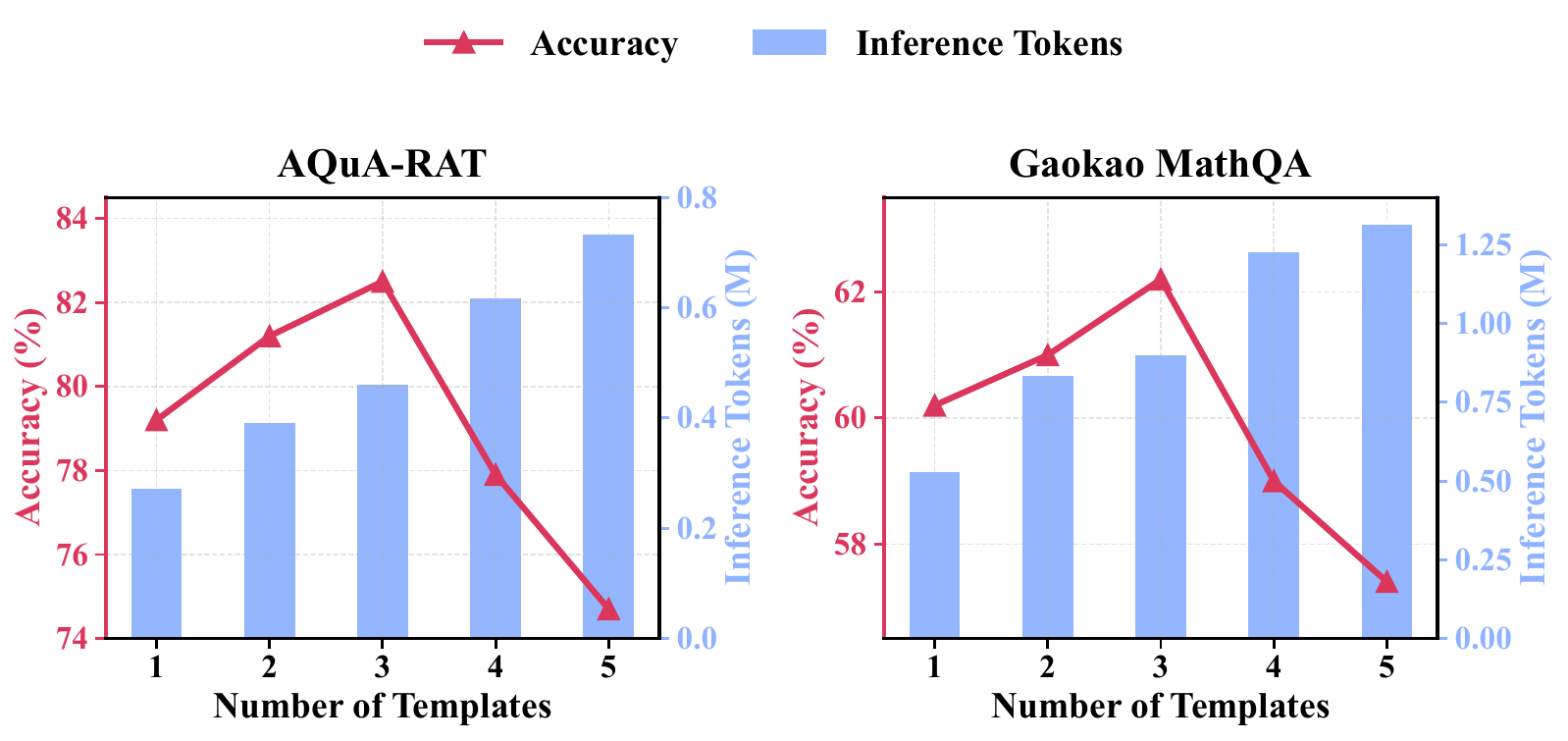}
  \caption{Ablation study on the impact of template number with GPT-4o-mini.}
  \label{fig:abl_num_templates}
\end{figure}
As shown in Figure ~\ref{fig:abl_num_templates}, we analyze the impact of Top-$k$ templates retrieved from CTM on model performance and inference token consumption on the mathematical domain using GPT-4o-mini. The performance curves on AQuA-RAT and Gaokao MathQA show similar trends: As the number of templates increases, the performance initially improves and then declines, peaking at the top 3 templates. A possible explanation is that a moderate number of templates offers effective reasoning guidance, whereas retrieving too many templates introduces redundant or noisy information and significantly increases the burden on the LLM to manage multiple templates, potentially leading to instability. In terms of inference token consumption, as the number of templates increases, the input context to the LLM becomes longer, resulting in a steady increase in token usage. Based on the above analysis, we select the top-$3$ templates to achieve a good balance of performance and costs.

\section{Conclusion}
In this paper, we revisit automatic prompt optimization from a new perspective. Instead of treating prompt optimization as a static prompt search problem, we formulate it as a continual experience accumulation process. Based on this insight, we propose \model, a self-evolving memory that enables LLM agents to progressively accumulate and reuse prompting knowledge across heterogeneous tasks. 
\model maintains a dual-memory mechanism that organizes successful reasoning strategies and failure signals into structured experiences, which are dynamically retrieved to guide prompt construction and refined through iterative self-reflection. Extensive experiments demonstrate that \model consistently improves performance while substantially reducing optimization cost. 

\section*{Limitations}
While \model demonstrates strong performance in automatic prompt optimization through self-evolving memory, some limitations remain. The current framework focuses on textual reasoning tasks and prompt optimization in language-based settings, without explicitly modeling multimodal information such as visual, audio, or structured environment signals. Extending the memory architecture to support multimodal experiences may further enhance the generality of the framework in broader agentic environments.

\bibliography{main} 

\appendix
\section{Appendix}
\subsection{Tasks and Data Details}
\label{appendix:tasks}
In this section, we present the details of the experimental dataset sizes and data splits in Table \ref{tab:dataset_split}. Table \ref{tab:dataset_license} presents the details of the dataset licenses and sources. All datasets are released under the MIT License or the Apache License 2.0 and are open for academic research. No personal privacy information is included. 

\begin{table}[h]
\centering
\caption{Dataset sizes and data splits.}
\resizebox{\columnwidth}{!}{
\begin{tabular}{lll}
\toprule
Dataset Name & Train \& Dev & Test \\
\midrule
BBH-Geometric Shapes & 150 & 100 \\
BBEH-Geometric Shapes & 100 & 100 \\
AGIEval-AQuA-RAT & 100 & 154 \\
AGIEval-Gaokao MathQA & 100 & 251 \\
AGIEval-Gaokao Geography & 100 & 99 \\
AGIEval-Gaokao History & 100 & 135 \\
\bottomrule
\end{tabular}
}
\label{tab:dataset_split}
\end{table}

\begin{table*}[ht]
\caption{License and Source of the datasets.}
\centering
\scalebox{0.6}{
\begin{tabular}{lll}
\toprule
Dataset                     & License & Source \\
\midrule
BBH-Geometric Shapes & MIT License & \url{https://github.com/suzgunmirac/BIG-Bench-Hard/blob/main/bbh/geometric_shapes.json}\\
BBEH-Geometric Shapes & Apache License 2.0 & \url{https://github.com/google-deepmind/bbeh/tree/main/bbeh/benchmark_tasks/bbeh_geometric_shapes} \\
AGIEval-AQuA-RAT & MIT License & \url{https://github.com/ruixiangcui/AGIEval/blob/main/data/v1/aqua-rat.jsonl}  \\
AGIEval-Gaokao MathQA & MIT License & \url{https://github.com/ruixiangcui/AGIEval/blob/main/data/v1/gaokao-mathqa.jsonl}  \\
AGIEval-Gaokao Geography & MIT License & \url{https://github.com/ruixiangcui/AGIEval/blob/main/data/v1/gaokao-geography.jsonl}  \\
AGIEval-Gaokao History & MIT License & \url{https://github.com/ruixiangcui/AGIEval/blob/main/data/v1/gaokao-history.jsonl}  \\
\bottomrule 
\end{tabular}
}
\label{tab:dataset_license}
\end{table*}

\paragraph{BBH-Geometric Shapes}
is a sub-task from the BIG-Bench Hard dataset \cite{bbh}. This task focuses on geometric reasoning, requiring the model to determine the geometric shape that would be generated based on the given SVG path element containing multiple commands. 
In the experiments, we follow the dataset split used by \cite{textgrad}.

\paragraph{BBEH-Geometric Shapes}
is a sub-task from the BIG-Bench Extra Hard dataset \cite{bbeh} that builds upon BBH by replacing each of the 23 tasks from BBH with a novel counterpart. BBEH-Geometric Shapes is much more difficult than the BBH-Geometric Shapes because each set of given commands can draw multiple shapes and may involve many useless commands. Following the \cite{textgrad}, we randomly split (seed=42) the data into training, validation, and test sets with a 50:50:100 ratio.

\paragraph{AGIEval-AQuA-RAT}
is a subset of the AGIEval benchmark \cite{zhong2024agieval}, focusing on evaluating the mathematical problem-solving ability using questions from the GRE and GMAT exams. In the experiments, 100 samples are randomly selected (seed=42) for the training and development set, with the remaining samples used for testing. AGIEval-Gaokao MathQA, AGIEval-Gaokao Geography, and AGIEval-Gaokao History adopt the same data split to facilitate the heterogeneous datasets' construction. For brevity, we omit repeating this split strategy when introducing these tasks later.

\paragraph{AGIEval-Gaokao MathQA}
is a subset of the AGIEval \cite{zhong2024agieval} that measures mathematical reasoning performance on Chinese college-entrance exam questions.

\paragraph{AGIEval-Gaokao Geography}
This AGIEval subset \cite{zhong2024agieval} targets geography-oriented question answering and is composed of problems collected from the Chinese Gaokao Geography exam.

\paragraph{AGIEval-Gaokao History}
Within AGIEval \cite{zhong2024agieval}, this subset focuses on historical understanding by evaluating responses to questions drawn from the Chinese Gaokao History exam.

\subsection{Baseline Details}
\label{appendix:baselines}
In this section, we report the configurations of the baselines used for comparison with \model in the experiments. Except for the analysis described in section \ref{sec:self_improving}, all LLM components serving as different roles in the methods are instantiated using the same underlying model.

\paragraph{ProTeGi} performs prompt optimization for 6 iterations with a minibatch size of 64. At each iteration, the prompt is evaluated on sampled training examples to identify errors, which are summarized into textual gradients describing how the prompt should be improved. The prompt is then edited in the opposite semantic direction of these gradients to generate new candidate prompts. A beam search with bandit-based selection retains the top 4 prompts for the next iteration.

\paragraph{OPRO} treats the large language model as an optimizer that iteratively generates candidate prompts based on a meta-prompt containing previously generated prompts and their performance scores. The iterative process runs for 20 rounds, with 8 prompts generated in each round.

\paragraph{TextGrad} conducts prompt optimization across 3 epochs with 4 steps per epoch, employing stochastic gradient descent with a batch size of 3. At each step, gradient signals are derived by back-propagating the feedback produced by the optimizer, enabling iterative refinement of the system prompt. A validation-based rollback mechanism is employed such that any update that reduces validation performance is discarded, reverting the prompt to its previous state.

\paragraph{PromptBreeder} optimizes prompts using an evolutionary algorithm that maintains a population of task-prompts and mutation-prompts. The evolution is initialized with 5 mutation prompts and 5 thinking styles, and runs for 20 generations. In each generation, the LLM generates mutated prompt variants, evaluates their performance (fitness) on training examples, and iteratively selects and evolves better prompts.

\paragraph{SPO} performs self-supervised prompt optimization without relying on external ground-truth labels. The optimization runs for 10 iterations, each iteration randomly selects 3 questions (without answers) from the training and validation datasets.

\subsection{Implementation Details}
The effectiveness of \model is validated on both a closed-source commercial model (GPT-4o-mini) and an open-source model (Qwen3-8B). 
For the dual-memory mechanism, we implement it on top of vector databases as the underlying storage layer. Embeddings are generated with Qwen3-Embedding-8B, and semantic similarity is utilized as the retrieval criterion. In Stage I, for CTM retrieval, we retrieve the top-3 templates during both training and inference, with similarity thresholds $\theta_{corr}$ set to 0.3 and 0.1, respectively; for EPM retrieval, we retrieve all recorded error patterns, as they represent the full spectrum of error types observed in previous attempts. In Stage II, for the self-reflection mechanism, the maximum number of retries is set to 3. In Stage III, for the CTM update, we introduce an additional verification step: each updated template's strategy must succeed on all 3 cases randomly sampled from the corresponding original correct template, otherwise the system will fall back into the template creation process. Moreover, the maximum number of templates in the system is capped at 30. Once this limit is exceeded, the template merge operation will be triggered to reduce template counts; for EPM update, the similarity threshold $\theta_{error}$ is set to 0.7. In the experiments we report, we conducted a single run.

\paragraph{Initial Prompt}
For the initial prompt, we use the original ones from \cite{bbh} for logical reasoning tasks, ``Let's solve the problem'' from \cite{tang2025unleashing} for mathematical calculation and knowledge-intensive tasks.

\subsection{Experimental Environments Details}
For all models used in this paper, we accessed them through their official APIs. The parameter sizes of closed-source models, such as GPT-4o-mini and GPT-5-Chat, are not publicly available. For the open-source setting, we employ an 8B large language model and an 8B embedding model, respectively. The total API compute budget was under one thousand US dollars. All experiments were conducted on a Linux-based system. The memory model implementation and reproduction are built based on the open-sourced vectore stores ChromaDB.

\subsection{More Experiments}
\paragraph{Computation Consumption}
\begin{table}[t]
\centering
\caption{Comparison of averaged API calls, token consumption, and optimization costs across different methods and backbones on Qwen3-8B and GPT-4o-mini.}
\label{tab:cost_analysis}
\scriptsize
\setlength{\tabcolsep}{6pt}
\renewcommand{\arraystretch}{1.05}

\resizebox{\columnwidth}{!}{
\begin{tabular}{l r r r}
\toprule
\textbf{Method} &
\textbf{Avg. Call$\downarrow$} &
\textbf{Avg. Token (M)$\downarrow$} &
\textbf{Avg. Cost(\$)$\downarrow$} \\
\midrule

\rowcolor{bbgreen}
\multicolumn{4}{c}{\textbf{Qwen3-8B} \qwenlogo} \\
ProTeGi        & 15682 & 17.14 & 7.44 \\
OPRO           &  4849 &  3.97 & 1.81 \\
TextGrad       &  2016 &  2.86 & 0.99 \\
PromptBreeder  & 27660 & 32.76 & 10.44 \\
SPO            & \textbf{75} & \textbf{0.37} & \textbf{0.08} \\
\hline
\textbf{\model}  & \underline{469} & \underline{1.60} & \underline{0.41} \\
\midrule

\rowcolor{bbblue}
\multicolumn{4}{c}{\textbf{GPT-4o-mini} \gptlogo} \\
ProTeGi        & 15451 & 13.27 & 4.52 \\
OPRO           &  2915 &  2.28 & 0.81 \\
TextGrad       &  2059 &  1.92 & 0.70 \\
PromptBreeder  & 27664 & 22.65 & 6.28 \\
SPO            & \textbf{75} & \textbf{0.30} & \textbf{0.05} \\
\hline
\textbf{\model}  & \underline{490} & \underline{1.39} & \underline{0.31} \\
\bottomrule
\end{tabular}
}

\end{table}
The specific values, including averaged API calls, token consumption, and optimization cost, are shown in table \ref{tab:cost_analysis}.

\subsection{Meta Prompt}
In this section, we present the meta prompts used in section \ref{sec:method} in Figure \ref{fig:meta_ans}, \ref{fig:meta_ref}, \ref{fig:meta_corr}, \ref{fig:meta_ctm_update}, \ref{fig:meta_merge}, \ref{fig:meta_sum}, \ref{fig:meta_error_update}.

\begin{figure*}[t] 
    \centering
\begin{tcolorbox}[
    enhanced,
    colframe=black,
    boxrule=0.5pt,
    title={\textcolor{white}{Answering Query Meta Prompt}},
    coltitle=white,
    fonttitle=\bfseries,
    attach boxed title to top left={xshift=2mm, yshift=-2mm},
    boxed title style={
        colback=black,
        sharp corners
    }
]
You are an expert assistant for question answering with retrieval-augmented memory.\par
Your job: answer the user's question by leveraging retrieved TEMPLATES as guidance and strictly following RULES derived from historical errors.\par

\vspace{2mm}
\{init\_instruction\}\par

\vspace{2mm}
\#\# RULES\par
The following rules are summarized from historical errors. You MUST follow them strictly:\par
\{rules\}\par 

\vspace{2mm}
<TEMPLATES>\par
Below are retrieved templates. Use their strategies as guidance. Each template includes one verified good case.\par
\{templates\}\par
</TEMPLATES>\par

\vspace{2mm}
<REFLECTIONS>\par
You have attempted this question before and failed. Learn from your mistakes and do NOT repeat them.\par
\{reflections\}\par
</REFLECTIONS>\par

\vspace{2mm}
<QUESTION>\par
\{question\}\par
</QUESTION>\par

\vspace{2mm}
<OUTPUT\_FORMAT>\par
\{output\_format\}\par
</OUTPUT\_FORMAT>\par
Before providing the final answer, outline your corresponding problem-solving steps, and avoid the mistakes mentioned in the rules. Finally, present your final answer according to the required output format.\par

\end{tcolorbox}
\caption{Meta prompt for answering query.} 
    \label{fig:meta_ans}
\end{figure*}

\begin{figure*}[t] 
    \centering
\begin{tcolorbox}[
    enhanced,
    colframe=black,
    boxrule=0.5pt,
    title={\textcolor{white}{Self Reflection Meta Prompt}},
    coltitle=white,
    fonttitle=\bfseries,
    attach boxed title to top left={xshift=2mm, yshift=-2mm},
    boxed title style={
        colback=black,
        sharp corners
    }
]
You are a precise self-reflection assistant.\par
Your job: analyze why the previous answer was wrong and extract a concise, actionable lesson.\par

\vspace{2mm}
Your previous answer to the following question is likely WRONG. Re-examine your reasoning and find the mistake.\par

\vspace{2mm}
<QUESTION>\par
\{question\}\par
</QUESTION>\par

\vspace{2mm}
<RULES>\par
The following rules are summarized from historical error\par
\{rules\}\par
</RULES>\par

\vspace{2mm}
<TEMPLATES>\par
Below are retrieved templates.\par
\{templates\}\par
</TEMPLATES>\par

\vspace{2mm}
<REFLECTIONs>\par
\{reflections\}\par
</REFLECTIONs>\par

\vspace{2mm}
<YOUR\_ANSWER>\par
\{wrong\_pred\}\par
</YOUR\_ANSWER>\par

\vspace{2mm}
Instructions:\par
1. Re-read the question carefully and check whether your answer actually addresses all constraints and conditions.\par
2. Trace through your reasoning step by step — identify any logical gaps, unjustified assumptions, or calculation errors.\par
3. Consider alternative interpretations or approaches you may have overlooked.\par
4. If prior reflections exist, do NOT repeat the same analysis — dig deeper or try a completely different angle.\par

\vspace{2mm}
You MUST respond with a JSON object in exactly this format and nothing else:\par
\{\par
    \quad "analysis": "your detailed step-by-step analysis of what went wrong", \par
    \quad "reflection": "one-sentence actionable lesson to avoid this mistake next time" \par
\}\par

\end{tcolorbox}
\caption{Meta prompt for self-reflection.} 
    \label{fig:meta_ref}
\end{figure*}

\begin{figure*}[t] 
    \centering
\begin{tcolorbox}[
    enhanced,
    colframe=black,
    boxrule=0.5pt,
    title={\textcolor{white}{Template Creation Meta Prompt}},
    coltitle=white,
    fonttitle=\bfseries,
    attach boxed title to top left={xshift=2mm, yshift=-2mm},
    boxed title style={
        colback=black,
        sharp corners
    }
]
You are an expert at abstracting reusable problem-solving templates.\par
Your job: given a question and its correct answer, extract a generalizable template that can guide solving similar problems in the future.\par

\vspace{2mm}
A model answered the following question correctly. Abstract this success into a reusable template.\par

\vspace{2mm}
<QUESTION>\par
\{question\}\par
</QUESTION>\par

\vspace{2mm}
<REFLECTIONs>\par
\{reflections\}\par
</REFLECTIONs>\par

\vspace{2mm}
<CORRECT\_ANSWER>\par
\{correct\_pred\}\par
</CORRECT\_ANSWER>\par
\{reflections\_block\}\par

\vspace{2mm}
Instructions:\par
1. Analyze the question type, scenario characteristics, and what makes this kind of problem recognizable.\par
2. Abstract the general solution procedure based on the successful reasoning trajectory. Describe the key reasoning or analysis steps needed to solve this type of problem. Each step should represent a high-level reasoning operation and can be directly reusable for a new problem that matches the same scenario. Keep the steps minimal, non-redundant, and sufficient for a single-pass solution attempt.\par
3. If prior failed attempts and reflections are provided, incorporate the lessons learned as pitfalls to avoid in the strategy.\par
4. Produce:\par
\quad - "when\_to\_use": a concise description of WHEN this template should be applied (what kind of question/scenario triggers it)\par
\quad - "strategy": the abstracted step-by-step reasoning procedure for this type of problem, including pitfalls to avoid if reflections are available\par

\vspace{2mm}
You MUST respond with a JSON object in exactly this format and nothing else:\par
\{\par
    \quad "when\_to\_use": "one-sentence description of the applicable scenario", \par
    \quad "strategy": "concise general reasoning strategy for this type of problem" \par
\}\par

\end{tcolorbox}
\caption{Meta prompt for template creation.} 
    \label{fig:meta_corr}
\end{figure*}

\begin{figure*}[t] 
    \centering
\begin{tcolorbox}[
    enhanced,
    colframe=black,
    boxrule=0.5pt,
    title={\textcolor{white}{Template Update Meta Prompt}},
    coltitle=white,
    fonttitle=\bfseries,
    attach boxed title to top left={xshift=2mm, yshift=-2mm},
    boxed title style={
        colback=black,
        sharp corners
    }
]
You are an expert template manager for a retrieval-augmented problem-solving system.\par
Your job: given a new successfully solved case and the recalled templates, decide the best action to keep the template library accurate, non-redundant, and maximally useful.\par

\vspace{2mm}
A model just answered a question correctly. Review the recalled templates and decide what actions to take.\par

\vspace{2mm}
<RECALLED\_TEMPLATES>\par
\{recalled\_templates\}\par
</RECALLED\_TEMPLATES>\par

\vspace{2mm}
<NEW\_GOOD\_CASE>\par
question: \{question\}\par
correct\_answer: \{correct\_pred\}\par
</NEW\_GOOD\_CASE>\par
\{reflections\_block\}\par
You must decide an action for EACH recalled template, and optionally add new templates. Rules:\par
- Each recalled template must appear EXACTLY ONCE in the actions list.\par
- Each action targets ONE template.\par

\vspace{2mm}
Available actions:\par
1. \textbf{none}: The recalled template already covers this case well (semantically equivalent). Keep it unchanged. Specify the template\_id.\par
2. \textbf{update}: The recalled template is relevant but its when\_to\_use or strategy can be enriched / made more comprehensive with information from the new case. Specify the template\_id and provide updated fields.\par
\quad - "when\_to\_use": a concise description of WHEN this template should be applied (what kind of question/scenario triggers it). Set to null to keep unchanged.\par
\quad - "strategy": the abstracted step-by-step reasoning procedure for this type of problem, including pitfalls to avoid if reflections are available. Set to null to keep unchanged.\par
3. \textbf{delete}: The recalled template conflicts with the new case (e.g. wrong strategy, contradictory advice) or is fully superseded. Specify the template\_id.\par
4. \textbf{add}: The new case represents a genuinely new problem type not covered by ANY recalled template. Create a new template. (Use sparingly — only when none of the recalled templates can be updated to cover this case.)\par
\quad - "when\_to\_use": a concise description of WHEN this template should be applied (what kind of question/scenario triggers it).\par
\quad - "strategy": the abstracted step-by-step reasoning procedure for this type of problem, including pitfalls to avoid if reflections are available.\par

\vspace{2mm}
IMPORTANT:\par
- Only use template\_id values that appear in RECALLED\_TEMPLATES above. Do NOT invent template\_ids.\par
- Every recalled template\_id must appear exactly once across all actions.\par

\vspace{2mm}
You MUST respond with a JSON object containing an "actions" list:\par
\{\par
    \quad "actions": [\par
        \quad\quad \{"action": "none", "template\_id": "..."\},\par
        \quad\quad \{"action": "update", "template\_id": "...", "when\_to\_use": "... or null", "strategy": "... or null"\},\par
        \quad\quad \{"action": "delete", "template\_id": "..."\},\par
        \quad\quad \{"action": "add", "when\_to\_use": "...", "strategy": "..."\}\par
        \quad ]\par
\}\par

\end{tcolorbox}
\caption{Meta prompt for template update.} 
    \label{fig:meta_ctm_update}
\end{figure*}

\begin{figure*}[t] 
    \centering
\begin{tcolorbox}[
    enhanced,
    colframe=black,
    boxrule=0.5pt,
    title={\textcolor{white}{Template Merge Meta Prompt}},
    coltitle=white,
    fonttitle=\bfseries,
    attach boxed title to top left={xshift=2mm, yshift=-2mm},
    boxed title style={
        colback=black,
        sharp corners
    }
]
You are an expert template librarian for a retrieval-augmented problem-solving system.\par
Your job: given a full list of templates, identify groups of templates that are semantically similar or overlapping and can be merged into a single, more general template without losing coverage.\par

\vspace{2mm}
The template library has grown too large (\{total\}) templates, limit is \{limit\}). Identify groups of templates that can be merged to reduce the total count.\par

\vspace{2mm}
<ALL\_TEMPLATES>\par
\{all\_templates\}\par
</ALL\_TEMPLATES>\par

\vspace{2mm}
Instructions:\par
1. Read ALL templates carefully. Identify groups where the when\_to\_use scenarios overlap significantly or the strategies are highly similar / complementary.\par
2. Only merge templates that are truly related --- do NOT force-merge unrelated templates just to reduce count.\par
3. For each merge group, produce a single merged template that covers all the scenarios and combines the best parts of each strategy.\par
4. Templates NOT included in any merge group will be kept as-is.\par
5. Try to reduce the total template count to at most \{target\} through merging.\par

\vspace{2mm}
You MUST respond with a JSON object in exactly this format and nothing else:\par
\{\par
\quad "merge\_groups": [\par
\quad\quad \{\par
\quad\quad\quad "template\_ids": ["1", "3"],\par
\quad\quad\quad "reason": "brief explanation of why these templates should be merged",\par
\quad\quad\quad "merged\_when\_to\_use": "combined scenario description covering all merged templates",\par
\quad\quad\quad "merged\_strategy": "combined strategy incorporating the best of each template"\par
\quad\quad \}\par
\quad ]\par
\}\par

\vspace{2mm}
If no templates can be reasonably merged, return: \{"merge\_groups": []\}\par

\end{tcolorbox}
\caption{Meta prompt for template merge.} 
    \label{fig:meta_merge}
\end{figure*}

\begin{figure*}[t] 
    \centering
\begin{tcolorbox}[
    enhanced,
    colframe=black,
    boxrule=0.5pt,
    title={\textcolor{white}{Error Summarization Meta Prompt}},
    coltitle=white,
    fonttitle=\bfseries,
    attach boxed title to top left={xshift=2mm, yshift=-2mm},
    boxed title style={
        colback=black,
        sharp corners
    }
]
You are an expert error-pattern analyst.\par
Your job: given a question, its correct answer, and multiple failed attempts with reflections, synthesize ONE concise, generalizable error-pattern description that can prevent similar mistakes in the future.\par

\vspace{2mm}
A model attempted the following question multiple times and FAILED every time. Analyze all attempts holistically and extract the root-cause error pattern.\par

\vspace{2mm}
<QUESTION>\par
\{question\}\par
</QUESTION>\par

\vspace{2mm}
<CORRECT\_ANSWER>\par
\{correct\_pred\}\par
</CORRECT\_ANSWER>\par

\vspace{2mm}
<FAILED\_ATTEMPTS>\par
\{failed\_attempts\}\par
</FAILED\_ATTEMPTS>\par

\vspace{2mm}
Instructions:\par
1. Compare all failed attempts — identify the COMMON root cause, not just surface-level symptoms.\par
2. Consider whether the errors stem from misunderstanding the question, flawed reasoning, calculation mistakes, or missing domain knowledge and so on.\par
3. Abstract the lesson into a generalizable rule that applies beyond this specific question.\par

\vspace{2mm}
You MUST respond with a JSON object in exactly this format and nothing else:\par
\{\par
    \quad "root\_cause": "brief description of the common root cause across all attempts", \par
    \quad "reflection": "one-sentence generalizable rule to prevent this category of error in the future" \par
\}\par

\end{tcolorbox}
\caption{Meta prompt for error summarization.} 
    \label{fig:meta_sum}
\end{figure*}

\begin{figure*}[t] 
    \centering
\begin{tcolorbox}[
    enhanced,
    colframe=black,
    boxrule=0.5pt,
    title={\textcolor{white}{Error Pattern Update Meta Prompt}},
    coltitle=white,
    fonttitle=\bfseries,
    attach boxed title to top left={xshift=2mm, yshift=-2mm},
    boxed title style={
        colback=black,
        sharp corners
    }
]
You are an expert error-pattern analyst.\par
Your job: refine an existing error-pattern description by incorporating new evidence from bad cases.\par
The updated pattern must be concise, generalizable, and actionable — it will be used as a RULE to prevent similar mistakes in the future.\par

\vspace{2mm}
An existing error pattern needs to be updated because a new bad case has been added to its cluster.\par

\vspace{2mm}
<CURRENT\_PATTERN>\par
\{current\_pattern\}\par
</CURRENT\_PATTERN>\par

\vspace{2mm}
<HISTORICAL\_BAD\_CASES>\par
The following are existing bad cases already in this error pattern cluster.\par
\{historical\_bad\_cases\}\par
</HISTORICAL\_BAD\_CASES>\par

\vspace{2mm}
<NEW\_BAD\_CASE>\par
This is the new bad case that triggered the update.\par
question: \{new\_question\}\par
correct\_answer: \{new\_ground\_truth\}\par
wrong\_answer: \{new\_wrong\_pred\}\par
reflection: \{new\_reflection\}\par
</FAILED\_ATTEMPTS>\par

\vspace{2mm}
Instructions:\par
1. Read the current pattern and historical bad cases to understand the existing error pattern.\par
2. Analyze the NEW bad case — determine whether it introduces a genuinely new dimension to the pattern.\par
3. If the new bad case is very similar to the historical ones and the current pattern already covers it well, you may keep the current pattern UNCHANGED.\par
4. Otherwise, produce a refined, generalizable one-sentence pattern description that:\par
\quad - Covers BOTH the new bad case and the historical ones\par
\quad - Is more precise or more general than the current pattern if the new evidence warrants it\par
\quad - Is actionable — it should clearly state what to do or avoid\par

\vspace{2mm}
You MUST respond with a JSON object in exactly this format and nothing else:\par
\{\par
    \quad "analysis": "brief reasoning about whether the new bad case changes or confirms the pattern", \par
    \quad "updated": "true or false (whether the pattern needs to be updated)"\par
    \quad "pattern": "one-sentence error-pattern description (refined if updated=true, or the original current pattern if updated=false)"\par
\}\par

\end{tcolorbox}
\caption{Meta prompt for error pattern update.} 
    \label{fig:meta_error_update}
\end{figure*}

\subsection{Ethical Statement}
\subsubsection{Scientific Artifacts}
All data sources and models underlying the experiment are used with explicit references and official links. Detailed information can be found in Appendix \ref{appendix:tasks}, \ref{appendix:baselines}. The experiments primarily target English and Chinese language questions across three domains: logical reasoning, mathematical problem solving, and knowledge-intensive question answering. The codes and experiment results will be released publicly with clear documentation, permitting use for research purposes. We manually collected the dataset according to its original requirements and, through manual review, excluded any data related to personal privacy.

\subsubsection{Potential Risks}
\model is designed for efficient and accurate prompt optimization, it not only improves task performance but also reduces optimization costs. However, potential risks may arise from unintended misuse, such as over-interpreting experiment results as indicators of real-world deployment readiness or being maliciously exploited to generate harmful or misleading content.

\subsection{Use of AI Assistants}
LLMs are used in this work strictly as auxiliary tools for limited language polishing of the manuscript, including improving fluency and presentation. All technical content, experimental analysis, and scientific claims are authored and finalized by the authors.

\end{document}